\documentclass[sigconf]{acmart}
\usepackage{colortbl}   
\usepackage{multirow}   
\usepackage{makecell}   
\usepackage{tabularx}
\AtBeginDocument{%
  }
\setcopyright{none}
\copyrightyear{2026}
\acmYear{2026}
\setcopyright{cc}
\setcctype{by}
\acmConference[KDD '26]{Proceedings of the 32nd ACM SIGKDD Conference on Knowledge Discovery and Data Mining V.2}{August 09--13, 2026}{Jeju Island, Republic of Korea}
\acmBooktitle{Proceedings of the 32nd ACM SIGKDD Conference on Knowledge Discovery and Data Mining V.2 (KDD '26), August 09--13, 2026, Jeju Island, Republic of Korea}
\acmDOI{10.1145/3770855.3818027}
\acmISBN{979-8-4007-2259-2/2026/08}
\begin{document}
\title{DREAM: LLM-based Dynamic Role-playing via Event-Aware Memory Graph}

\author{Zhihao Xiao}
\authornote{Both authors contributed equally to this research.}
\orcid{0009-0009-5779-1345}
\affiliation{%
  \institution{Hangzhou International Innovation Institute, Beihang University}
  \city{Hangzhou}
  \country{China}
}
\email{xiaozhihao@buaa.edu.cn}

\author{Mengting Li}
\authornotemark[1]
\affiliation{%
  \institution{Hangzhou International Innovation Institute, Beihang University}
  \city{Hangzhou}
  \country{China}
}
\email{li_mengting@buaa.edu.cn}

\author{Xintao Wang}
\affiliation{%
  \institution{School of Computer Science, Fudan University}
  \city{Shanghai}
  \country{China}
}
\email{xtwang21@m.fudan.edu.cn}

\author{Linfeng Li}
\affiliation{%
  \institution{Hangzhou International Innovation Institute, Beihang University}
  \city{Hangzhou}
  \country{China}
}
\email{zy2457211@buaa.edu.cn}

\author{Limin Shui}
\affiliation{%
  \institution{Hangzhou International Innovation Institute, Beihang University}
  \city{Hangzhou}
  \country{China}
}
\email{Liminshui@buaa.edu.cn}

\author{Mengqi Ji}
\affiliation{%
  \institution{Hangzhou International Innovation Institute, Beihang University}
  \city{Hangzhou}
  \country{China}
}
\email{jimengqi@buaa.edu.cn}

\author{Borui Cai}
\authornote{Corresponding author.}
\affiliation{%
  \institution{Hangzhou International Innovation Institute, Beihang University}
  \city{Hangzhou}
  \country{China}
}
\email{caibr@buaa.edu.cn}

\begin{abstract}
Role-playing agents (RPAs) have emerged as a key application of large language models, enabling immersive and high-fidelity character simulation. Accurate role-playing of established characters requires not only stylistic imitation but also temporally consistent and causally grounded behavioral reasoning. However, existing RPAs primarily rely on static character descriptions and unstructured memory, limiting their ability to maintain long-term narrative and personality coherence. We introduce DREAM, a structured memory framework for role-playing agents inspired by the Activating Event–Belief–Consequence (ABC) cognitive model. DREAM transforms unstructured literary text into an Event-aware Memory Graph (EMG) that organizes character experiences into temporally ordered and causally linked event graph. This representation enables the construction of dynamic, dual-granularity character profiles that capture both stable personality traits and event-driven behavioral evolution. We further propose the Temporal Causal Memory (TCM) benchmark to evaluate temporal consistency and long-range causal narrative coherence. DREAM achieves state-of-the-art performance across CoSER, LIFECHOICE, and TCM, outperforming multiple strong baselines. Our approach demonstrates the effectiveness of structured memory in enhancing the interpretability and consistency of role-playing agents.
\end{abstract}




\maketitle

\section{Introduction}
\begin{figure}
  \includegraphics[width=\linewidth]{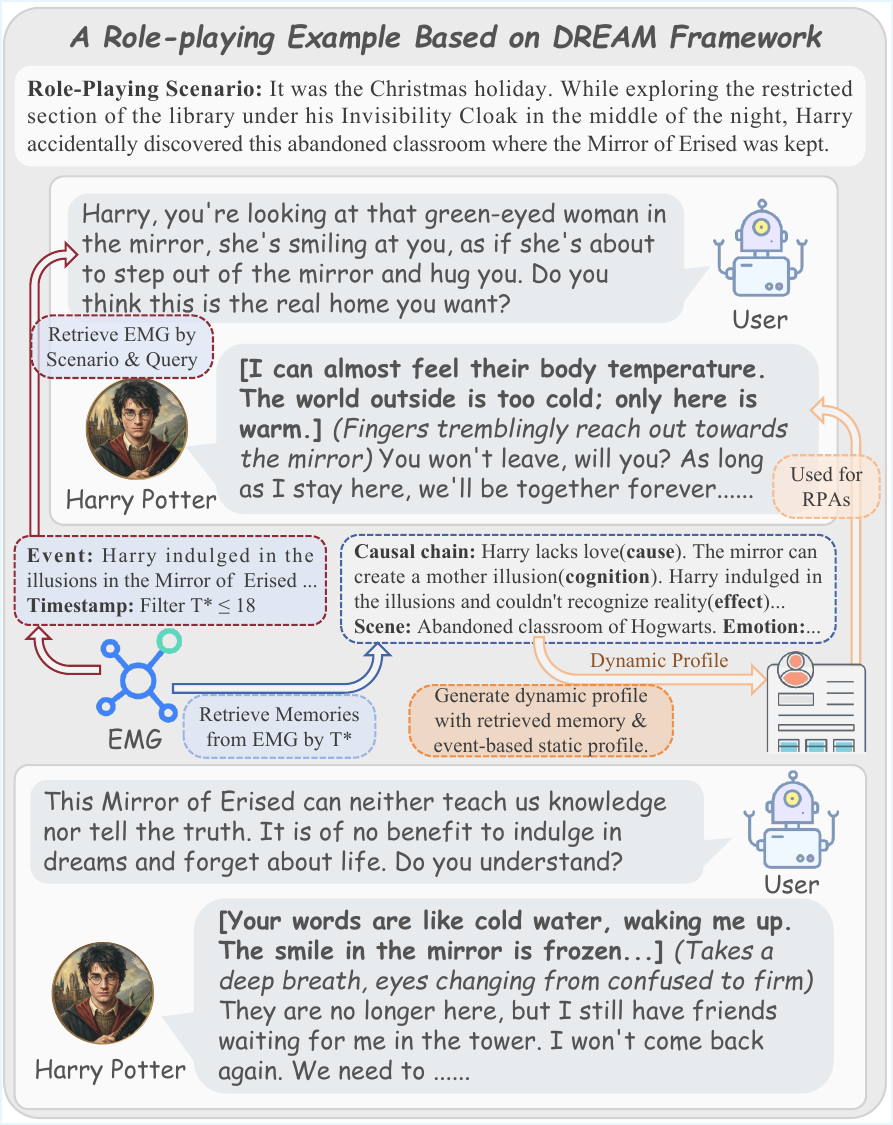}
  \caption{A role-playing example of the DREAM framework. The framework leverages the EMG and temporal filtering to construct dynamic profiles, enabling LLMs to generate contextually accurate and emotionally resonant interactions, such as Harry Potter’s encounter with the Mirror of Erised.}
  \label{fig:example_chat}
\end{figure}

Recent advances in large language models (LLMs) have enabled a new generation of role-playing agents (RPAs) that can simulate the linguistic styles and reasoning patterns of specific characters~\citep{tseng2024two, chen2024persona}. Such agents have shown promise in immersive games~\citep{xu2025empowering}, interactive fiction, and virtual companions, where maintaining role consistency over long and dynamic interactions is critical~\citep{park2023generative, zhou2023sotopia}. However, despite recent progress~\citep{zhou2024characterglm, wang2024rolellm}, constructing RPAs for established characters remains challenging. Beyond surface-level persona imitation, effective role-playing requires agents to align their responses with a character's past experiences, cognition, and motivations as shaped by the underlying narrative over time, as shown in Fig.~\ref{fig:example_chat}.

RPAs typically rely on two kinds of persona data: profiles and memories. However, existing approaches exhibit two key limitations in effectively leveraging resources: (1)\textbf{Static Profiles:} Most existing methods inject predefined character profiles designed for fixed scenarios through prompting or fine-tuning~\citep{lu2024large, zhou2024characterglm, he2025crab}. While effective at capturing surface-level stylistic traits, such representations fail to reflect how a character’s disposition is shaped and updated by past experiences. As a result, RPAs often respond rigidly and struggle to adapt to dynamic interactions, where the profile of a character should be dynamically grounded in its memory rather than predefined scenarios. (2)\textbf{Fragmented Memory:} Prior work on character memory typically follows two paradigms: retrieving past dialogues or events based on semantic similarity~\citep{li2023chatharuhi, xu2025character}, or encoding experiences implicitly within model parameters~\citep{wang2025coser}. Although these approaches enable memory access, they organize character memory as isolated units, thereby fragmenting complete plots, leaving temporal order and inter-event causal relationships implicit. Consequently, RPAs can recall what happened, but cannot reliably reason about when and why past experiences should influence current behavior, leading to narrative inconsistency and temporal leakage~\citep{ahn2024timechara}. Moreover, parametric memories often discard fine-grained event details and offer limited interpretability, further constraining their applicability in character-centric reasoning.

To address these challenges, we propose DREAM (\textbf{D}ynamic \textbf{R}ole-playing via \textbf{E}vent-\textbf{A}ware \textbf{M}emory), a role-playing multi-agent system with event-aware memory and dynamic profiles. DREAM explicitly organizes character experiences as an Event-aware Memory Graph (EMG), where events are organized by temporal order and causal dependencies. In EMG, each event encodes both global narrative context and fine-grained character dynamics, enabling agents to reason over how past experiences shape character development. During interaction, DREAM retrieves temporally valid memories from the EMG to dynamically synthesize event-based character profiles, ensuring response generation remains consistent with the character's time-dependent cognitive state in the storyline.

Specifically, DREAM constructs the EMG by extracting information from literary texts at dual granularities, i.e., global narrative context at the macro level and fine-grained dynamics at the micro level. At the macro level, it captures event background, world settings, and character attributes. At the micro level, it adopts causal narrative chains to capture fine-grained emotion-cognition-behavior dynamics within unit plots, inspired by the well-established psychological Activating Event-Belief-Consequence (ABC) model~\citep{ellis1957rational}. During role-playing, DREAM employs a temporally constrained hybrid retrieval that integrates semantic matching with graph-based multi-hop inference; this mechanism enables the agent to trace the causal origins of a character’s behavior and evolution, ensuring that responses are grounded in a logically coherent and chronologically valid narrative context.

To complement evaluation dimensions that are underexplored in prior work, we propose the TCM benchmark, which focuses on two key aspects: (1) coherent temporal memory and (2) long-term causal narrative consistency. We conduct comprehensive experiments on CoSER, LIFECHOICE, and TCM, and the results show that DREAM achieves state-of-the-art performance across baselines.

Our contributions are summarized as follows:
\begin{itemize}
\item We introduce DREAM, a multi-agent framework for role-playing that systematically constructs memory graphs and generates dynamic character profiles to support context-aware interactions.
\item At the core of DREAM, we propose the Event-aware Memory Graph (EMG), which encodes temporally ordered events and causal dependencies, enabling dynamic profile construction and contextually grounded retrieval.
\item Extensive experiments on various benchmarks demonstrate that DREAM achieves state-of-the-art performance across baselines, consistently improving temporal narrative consistency and role-playing ability.
\end{itemize}

\section{Related Work}
\subsection{Role-Playing Agents with LLMs}
Role-playing agents (RPAs) aim to achieve high-fidelity simulations of specific personas. Prior research has mainly focused on: (1) \textbf{Profile Construction:} Injecting static profiles via SFT or prompting. \textbf{Ditto} \citep{lu2024large} utilizes self-alignment mechanisms to elicit intrinsic character knowledge from LLMs, whereas \textbf{CharacterGLM} \citep{zhou2024characterglm} integrates diverse sources including human role-playing and literature extraction. To further enhance persona depth, profiles are often enriched with psychological attributes \citep{occhipinti2024prodigy} or multimodal information \citep{dai2024mmrole, zhang2025omnicharacter}. Moreover, \textbf{Crab} \citep{he2025crab} introduces a configurable framework to improve profile flexibility. (2) \textbf{Cognitive Alignment:} Constraining the model's internal reasoning to align its cognitive mechanisms with the target character. Tang et al. \citep{tang2025thinking} employ character-centric Chain-of-Thought (CoT) to explicitly guide reasoning paths. Alternatively, reinforcement learning approaches internalize motivations through contrastive preferences \citep{ye2025cpo}, dual-process architectures \citep{liu2025cogdual}, or multi-objective alignment \citep{liao2025moa}. However, existing studies predominantly rely on static character profiles, limiting their ability to adapt behaviors across evolving interaction contexts. This limitation motivates research on incorporating event-aware memory mechanisms into RPAs.

\subsection{Agent Memory Systems}
Memory systems for LLM-based agents~\citep{zhang2025survey, hu2025memory} primarily rely on parametric storage within model weights~\citep{fang2024alphaedit} or external retrieval paradigms~\citep{yan2025memory}. Notably, GraphRAG~\citep{edge2024local} advances the latter by leveraging graph-based indexing to capture both local and global semantic relationships. Recently, specialized memory mechanisms for RPAs have emerged~\citep{chen2023large,wang2024rolellm}. For instance, ChatHaruhi~\citep{li2023chatharuhi} retrieves relevant past dialogs via similarity-based retrieval, which primarily captures conversational style. CHARMAP~\citep{xu2025character} stores key events in a key-value memory bank, yet its chunk-based extraction may fragment complete plots. CoSER~\citep{wang2025coser} further incorporates character motivations at the event level, but it overlooks the causal dependencies between those events. Overall, these methods suffer from fragmented memory organization, limited interpretability, and a lack of explicit inter-event causal structure, resulting in largely static character representations. In contrast, we construct explicit character memories at the event granularity, organizing experiences into structured causal chains to support interpretable and consistent long-term role-playing.

\section{The DREAM Framework}
In this section, we elaborate on the design of DREAM. 
The primary goal of DREAM is to support dynamic role-playing through the EMG, with two key objectives:

\begin{itemize}
\item Establish the EMG as a structured memory that encodes the character’s experiences into causally linked event chains, facilitating a holistic understanding of their behavioral logic.
\item Instantiate dynamic character profiles through EMG-based retrieval, effectively mapping the character's time-dependent cognitive state throughout the interaction.
\end{itemize}

The design of DREAM stems from a requirement that high-quality RPAs must possess memory continuity and narrative depth. DREAM follows a two-stage pipeline: first, it transforms unstructured literary texts into the structured EMG that encodes temporal and causal dependencies. During role-playing, it performs targeted retrieval from this EMG to instantiate dynamic, context-aware character profiles for response generation.

\subsection{Core Formulation of DREAM}
DREAM enables dynamic and experience-driven role-playing by maintaining character memory with explicit temporal and causal structure. This capability is supported by the Event-Aware Memory Graph (EMG), which organizes fragmented character experiences into a structured memory graph with timestamps and causal dependencies.

\subsubsection{Problem Formulation.}
In a literary role-playing setting, given a literary corpus $\mathcal{B}$ and a target character $c$, our aim is to optimize LLMs to simulate the behavior and personality $P$ of $c$. We define a character as a dual-structured representation:
\begin{equation}
\mathcal{P}_c=(\text{ }\mathcal{A}\text{ },\mathcal{G}_{event}\text{ })
\end{equation}
where $\mathcal{A}$ denotes the identity attributes of the character (e.g., background, social roles, and inherent personality traits), and $\mathcal{G}_{event}$ represents an EMG. Unlike traditional flat memory banks, $\mathcal{G}_{event}$ is a structured representation extracted from original text that encodes character persona, experiences, causal dependencies, and psychological evolution.

Grounding interactions in $\mathcal{G}_{event}$, the DREAM framework follows a retrieve-generate-respond workflow. At each dialogue turn, we utilize the user query $Q$ as a probe to traverse $\mathcal{G}_{event}$ and retrieve: (1) anchor event, the event most relevant to $Q$ and the scenario $S$; and (2) causal chains, which represent the causal narrative chains encapsulating character cognition and motivations within anchor events. DREAM performs a two-stage process to generate a dynamic profile:
\begin{itemize}
\item Event Information Retrieval ($\mathcal{R}_{event}$): Retrieve the event-based character personalities that most fit the current scenario by traversing the timestamps $t_i$.
\item Dynamic Profile Instantiation ($\mathcal{P}_{dyn}$): Based on the retrieved context, we model the character's time-dependent cognitive state, belief updates, and personality evolution.
\end{itemize}
The process is formalized as follows:

\begin{equation}
\label{eq:profile_formulation}
\mathcal{P}_{dynamic} = \text{Generate}(\text{R}_{event}(Q, S))
\end{equation}
Before each response is generated by the LLM (Eq.~\eqref{eq:respond_model}), DREAM adopts the above generated dynamic profile to appropriately respond to user query: 
\begin{equation}
\label{eq:respond_model}
Respond = LLM(Q, \mathcal{R}_{event}, \mathcal{P}_{dynamic})
\end{equation}

\subsubsection{The EMG Definition}
The design of EMG aims to overcome the limitations of static profiles and fragmented memory of existing role-playing agents (RPAs):

EMG organizes memories with event $E_i$ as the basic unit and connects experiences at different stages through explicit temporal and causal structures. Then, retrieved context from the EMG can form dynamic profiles that accurately reflect the cognitive states, motivation updates, and personality evolution, thereby moving beyond surface-level stylistic imitation. To achieve that, we define EMG as a multi-attribute-directed knowledge graph:
\begin{equation}
\mathcal{G}_{event} = (E_{macro,micro} , R_{temporal,causal})
\end{equation}

Events are central entities in EMG, and we also design peripheral entities to represent detailed information of entities at macro and micro levels $E_{macro,micro}$. The macro-level includes entities that reflect event background information (Worldview, Environments) and character's current status (Identity, Social Relationships, Appearance, Habitual Actions, and Representative Dialog). The micro-level includes entities that depict fine-grained narrative and psychological units (Unit-Plots, Emotions, Cognition, Behaviors, Specific Scenes, Items, and Skills).

We organize relations in EMG into two types $R_{temporal,causal}$: temporal relations and causal relations. Temporal relations connect events along the narrative timeline (e.g., $next\_event, next\_plot$), ensuring coherent chronological memory organization.
Causal relations capture dependencies among character cognition, emotions, behaviors, and event factors (e.g., $motivated\_by$, $emotion\_from$), modeling transitions across internal states and external actions. In addition, EMG includes a small set of conventional relations to represent standard entity relationships. Formal definitions of entities and relations are available in the Appendix~\ref{sec:KG_appendix}.

\subsection{EMG Construction}
We develop a Memory Construction Agent to construct EMG from unstructured literary texts, such as novels and plays. The aim is to explicitly encode temporally ordered causal events and rich character information, which are later used to derive event-based dynamic character profiles.

\subsubsection{Event and Temporal Structure Construction.}
Event Construction is performed through chapter-level narrative segmentation followed by character-centric event categorization. 
We first segment literary texts at the chapter level and employ LLMs to identify chapter boundaries and remove non-narrative sections, which is adopted to preserve event completeness and mitigate plot fragmentation introduced by conventional fixed-length or sentence-level segmentation methods.

Each chapter is then categorized based on character relevance and event completeness into complete events, incomplete events, or contextual content. \textbf{Complete event:} An event including the cause, process, and outcome of the character. \textbf{Incomplete event:} Chapter related to the character but insufficient to form a complete event. In this case, next chapter is appended until a complete event can be identified. \textbf{Contextual content:} Content that does not involve any character events and can serve as supplementary information, and is incorporated into the EMG.

To enable temporally aware retrieval and prevent future leakage (i.e., spoilers), we add a $t_i$ to every $E_i$ and $U_p$ node based on the extracted narrative sequence. This ensures that RPAs can only access memories preceding the current narrative time $t_i$ during role-playing.

\begin{figure}[t]  
  \centering
  \includegraphics[width=\linewidth]{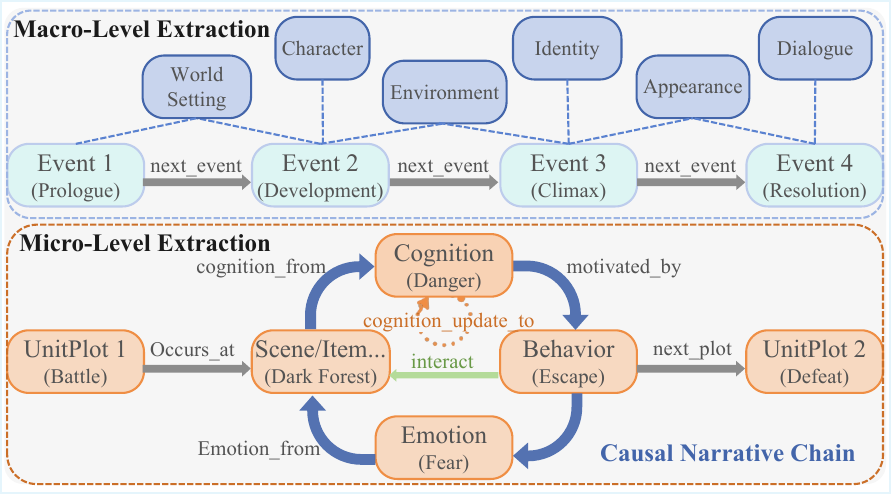}
  \caption{The structure and definition of knowledge graph}
  \label{fig:graph_structure}
\end{figure}
\begin{figure*}
  \centering       
  \includegraphics[width=\textwidth]{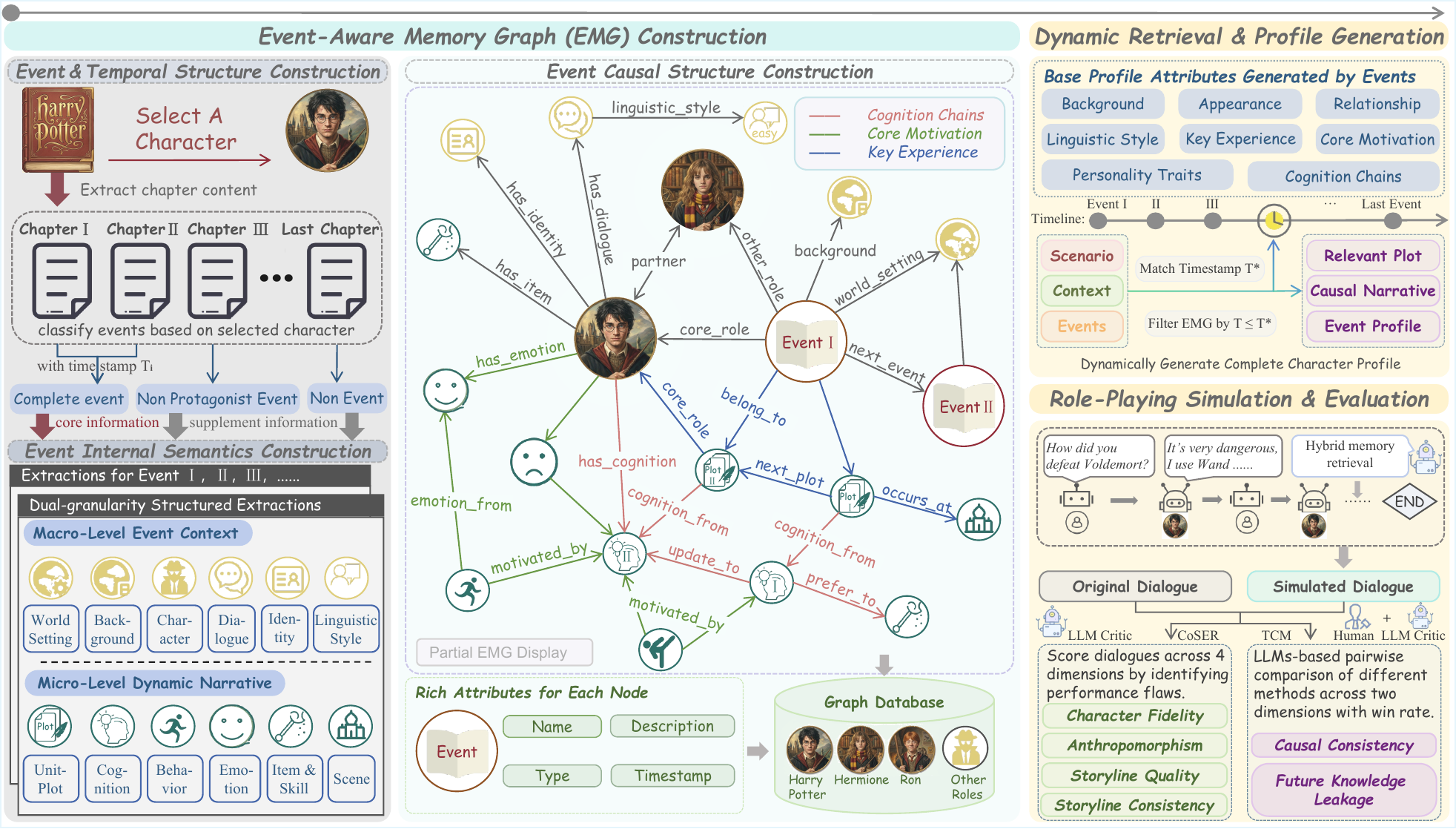}
  \caption{The complete pipeline of DREAM. Left: DREAM is sourced from renowned books and processed via an LLM-based pipeline. By classifying the events of the selected character and performing dual-granularity structured extraction, entities are obtained. Middle: Macro-level (yellow) and micro-level (green) nodes are integrated to construct the EMG containing narrative chains. Right: Timestamp filtering is used to dynamically generate character profiles, which are then used in dialogue simulations and evaluated from multiple dimensions by LLMs and humans.}
  \label{fig:framework}
\end{figure*}

\subsubsection{Event Internal Semantics Construction.}\label{sec:info_extract}
Existing extraction approaches, such as dialogue-centric extraction~\citep{li2023chatharuhi} and chunk-level extraction~\citep{xu2025character}, fail to jointly model global event structure and fine-grained character dynamics over time.
To construct a structured memory framework capable of simultaneously capturing personas and the evolution of $C_o$ and experiences in literary works, we propose dual-granularity knowledge extraction, as shown in Fig.~\ref{fig:graph_structure}:
\paragraph{Macro-Level Event Context} This level operates at an event-level abstraction, capturing world settings, background information, and character attributes for the $i$-th event, denoted as $E_i$. This structured context enables agents to maintain narrative consistency when reasoning about specific events. We further incorporate a timestamp $t_i$ into the event sequence to prevent future-event leakage during memory retrieval. For each $E_i$, we define entities and relations (partially displayed in Fig.~\ref{fig:framework}, details in Table~\ref{tab:macro_def}) to extract a structured tuple of contextual attributes.
\paragraph{Micro-Level Dynamic Narrative} To capture the causal logic behind character development, we introduce fine-grained $U_p$ extraction. Specifically, we define $U_p$ as a fine-grained sub-unit within $E_i$. Rather than simply recording what happened, we model why it happened through chains of emotion, cognition, and behavior in each associated $U_p$. The extraction captures character development, emotional and cognitive changes, and behavioral causal relationships within each event. For each event, we define entities and relations to extract a set of structured attributes and define the narrative flow as a directed graph of $U_p$. 

For coreference resolution during graph construction (such as the same \texttt{Emotion} in $E_i$), we ensure consistent entity naming via prompt-guided extraction. (Detailed entity and relation definitions, along with extraction prompts, are provided in Appendix~\ref{sec:KG_appendix}.)


To ensure a coherent narrative structure, we employ an entity resolution strategy that automatically consolidates recurring entities across various event segments. This maintains the continuity of character identities and integrates disparate event nodes into a unified, timeline-based knowledge graph.

\subsubsection{Event Causal Structure Construction.}\label{sec:graph_construction}
To ensure consistent character memory representation and efficient retrieval, we propose a construction pipeline. Following chronological order with roles and events as core nodes, we combine the extracted nodes into triples through predefined relationships on an event-by-event basis and add them to the graph until the complete EMG is built. We employ a synchronized data ingestion strategy to efficiently fuse structured information from both macro- and micro-levels into the EMG (Fig.~\ref{fig:framework}).

In complex stories, characters do not act randomly; their behaviors are motivated by perceptions and experiences. To integrate this cycle of experience, cognition, and action into the EMG, we design a causal narrative construction mechanism:

\textbf{Cognitive Chain:} This chain models the dynamic evolution of a character's mental state throughout the narrative. It is achieved by defining entities (e.g., \texttt{cognition}, \texttt{behavior}) and relations (e.g., \texttt{motivated-by}, \texttt{cognition-update-to}) to accurately capture how cognitive states drive behavioral changes. The red lines of the EMG in Fig.~\ref{fig:framework} show part of the Cognition Chains.

\textbf{Causal Chain:} The causal chain reveals character development. By selecting nodes (e.g., \texttt{Unit-Plot}, \texttt{Behavior}) and relations (e.g., \texttt{next-plot}, \texttt{motivated-by}), we construct a causal chain based on the character's motivation and experience. This chain explicitly represents the inherent causal dependencies between character behaviors and events, providing a robust structured foundation for scenarios such as plot and behavior generation. The green and blue lines of the EMG in Fig.~\ref{fig:framework} form a causal narrative chain.

\subsection{Memory Retrieval and Dynamic Profile Generation} \label{sec:memory_retrieval}
This section introduces how DREAM utilizes a temporal memory retrieval mechanism to dynamically generate profiles for role-playing. Based on the constructed EMG, we first retrieve the subgraphs corresponding to 8-dimensional (shown in Fig.~\ref{fig:framework}) personas and generate descriptions of personas via the subgraphs.

Through given scenarios or contexts during the role-playing, we retrieve matched event nodes and their corresponding timestamp. Before RPAs generate responses, DREAM rewrites connected causal narrative subgraphs into descriptive paragraphs, so as to realize the dynamic update of profile during the role-playing process.
\subsubsection{Hybrid Memory Retrieval Mechanism}\label{retri}
To ensure semantic consistency, narrative logical rigor, and retrieval efficiency, while mitigating hallucinations of LLMs during long-range role-playing, we propose a hybrid retrieval architecture:

\paragraph{Semantic Similarity Retrieval} To retrieve relevant event information, we utilize cosine similarity to perform semantic matching between the descriptions of \texttt{Event} and the current context. This process enables DREAM to efficiently locate  top-$K$ most relevant events. And we obtain the biggest $t_i$ in these events for temporal retrieval.

\paragraph{Time-Constrained Memory Retrieval} To retrieve Character trait information and uphold strict narrative logic with preventing future leakage, we implement a temporal memory retrieval mechanism. This mechanism restricts the searchable memory space to events occurring before the current timestamp. The mechanism is applied to retrieve events and specific nodes—such as \texttt{Unit-Plot}, and \texttt{Cognition}—ensuring all retrieved details are chronologically valid for the character’s current state.

\paragraph{Structure-based Multi-hop Retrieval} To capture character's long-range causal dependencies or the abstract evolution of cognition. We leverage the graph structure of the EMG for deep memory retrieval:

\textbf{Cross-dimensional Retrieval:} By traversing relations such as \texttt{motivated-by} and \texttt{cognition-from}, the agent can trace the underlying psychological drivers and cognitive shifts behind a character's behavior.

\textbf{Causal Chain Retrieval:} When query involves a character’s long developmental history or complex relationships, we implement a multi-hop retrieval strategy, which process initiates from \texttt{Unit-Plot} and propagates along edges to harvest connected \texttt{Cognition} and 
\texttt{Behavior} nodes. This allows agents to synthesize details into a coherent causal chain (e.g., linking a past psychological trauma to a current reaction), resulting in dialogue responses with greater personality depth.

\begin{table*}
    \caption{The performance(\%) of DREAM and baselines on the comprehensive role-playing benchmark CoSER. \underline{Underlined} values indicate best performance across each model, \textbf{Bold} values indicate best performance in all models.}
    \label{tab:coser_results}
        \begin{tabular}{l|l|cccc|c}
            \toprule
            \textbf{ Models} & \textbf{Methods} & \makecell{\textbf{Storyline} \\ \textbf{Consistency}} & \textbf{Anthropomorphism} & \makecell{\textbf{Character} \\ \textbf{Fidelity}} & \makecell{\textbf{Storyline} \\ \textbf{Quality}} & \textbf{ Average } \\
            \midrule
            \rowcolor{gray!15}
            \multicolumn{7}{c}{\textbf{Closed-Source LLMs}} \\
            \midrule
            \multirow{4}{*}{GPT-4o}
            & RAG      & 56.33 & 42.96 & 42.54 & 67.16 & 52.50 \\
            & GraphRAG & 59.34 & 45.62 & 43.71 & 71.16 & 54.96 \\
            & GCA      & 61.83 & \underline{\textbf{49.66}} & 47.75 & 78.16 & 59.35 \\
            & DREAM    & \underline{\textbf{67.25}} & 49.36 & \underline{50.45} & \underline{\textbf{84.38}} & \underline{\textbf{62.86}} \\
            \midrule 
            \multirow{4}{*}{Gemini-2.5-Pro} 
            & RAG      & 58.70 & 43.79 & 45.74 & 69.26 & 54.37 \\
            & GraphRAG & 56.54 & \underline{46.22} & 46.17 & 68.83 & 54.44 \\
            & GCA      & 59.85 & 45.83 & 46.95 & 78.86 & 57.87 \\
            & DREAM    & \underline{63.25} & 43.36 & \underline{49.75} & \underline{81.38} & \underline{59.44} \\
            \bottomrule
            
            \rowcolor{gray!15} 
            \multicolumn{7}{c}{\textbf{Open-Source LLMs}} \\
            \midrule
            \multirow{4}{*}{LLaMA-3.1-8B} 
            & RAG      & 49.77 & 40.75 & 35.94 & 59.76 & 46.56 \\
            & GraphRAG & 50.34 & 41.62 & 35.37 & 61.29 & 47.16 \\
            & GCA & 53.63 & 45.56 & 40.75 & 72.86 & 53.20 \\
            & DREAM& \underline{58.25} & \underline{45.85} & \underline{45.95} & \underline{78.83} & \underline{57.22} \\
            \midrule 
            \multirow{4}{*}{Qwen-2.5-72B} 
            & RAG      & 53.77 & 45.73 & 45.54 & 68.76 & 53.45 \\
            & GraphRAG & 55.34 & 45.42 & 45.12 & 70.29 & 54.04 \\
            & GCA      & 58.72 & 48.08 & \underline{\textbf{51.49}} & 74.10 & 58.10 \\
            & DREAM    & \underline{60.36} & \underline{49.96} & 49.35 & \underline{79.38} & \underline{59.76} \\
            \bottomrule
                    
            \rowcolor{gray!15} 
            \multicolumn{7}{c}{\textbf{Role-Playing LLMs}} \\
            \midrule
            \multirow{4}{*}{CharacterGLM-6B} 
            & RAG & 48.36 & 41.68 & 36.70 & 63.86 & 47.65 \\
            & GraphRAG & 50.74 & 43.58 & 39.34 & 66.21 & 49.97 \\
            & GCA & 53.33 & \underline{45.66} & 41.50 & 69.14 & 52.41 \\
            & DREAM & \underline{54.95} & 45.46 & \underline{42.45} & \underline{72.28} & \underline{53.79} \\
            \midrule
            \multirow{4}{*}{CoSER-8B} 
            & RAG & 51.64 & 44.68 & 39.78 & 70.50 & 51.65 \\
            & GraphRAG & 55.94 & 45.96 & 42.98 & 71.29 & 54.04 \\
            & GCA & 58.73 & \underline{48.56} & \underline{46.55} & 73.45 & 56.82 \\
            & DREAM & \underline{60.45} & 47.36 & 46.15 & \underline{78.28} & \underline{58.06} \\
        
            \bottomrule
      \end{tabular}
\end{table*}

\subsubsection{Dynamic Profile Generation}
Considering the speed and accuracy of profile generation during dynamic role-playing, we propose the dynamic profile constructed by rewriting the EMG information based on dynamic retrieval.

\textbf{Event-based Information}: We provide an automatic generation method based on incremental updating inspired by Hollmwood~\citep{chen2024hollmwood}. We employ LLMs to automatically synthesize multi-dimensional character descriptions derived from the EMG to construct event-based character personalities, including background, appearance, linguistic style, key experience, core motivation, relationship, cognition chains and personality traits. 

\textbf{Dynamic Profile}: For the context or a given scenario in the current role-playing, DREAM uses semantic similarity retrieval to obtain the anchor events with $t_i$ and retrieves the memory subgraph through hybrid retrieval (see \ref{retri}), and rewrites the parts of the $\mathcal{R}_{event}$ that need to be updated via LLMs. Finally, the profile that is most suitable for the current context is dynamically generated.

\begin{table}[t!]
    \caption{Results of 3 LLMs with 3 methods on LIFECHOICE. ACC refers to the decision accuracy. +motivation refers to the results with character motivations.}
    \label{tab:lifechoice_results}
        \begin{tabular}{llccc}
            \toprule
            \textbf{Method} & \textbf{Model} & \textbf{ACC} & \textbf{+motivation} \\
            \midrule
            
            \rowcolor{gray!15}
            \multicolumn{4}{c}{\textbf{Profile \& Memory}} \\
            \midrule
            
            \multirow{3}{*}{GraphRAG} 
            & GPT-4o  & 68.15 & 94.25 \\
             & LLaMA-3.1-8B  & 61.22 & 92.43 \\
             & CoSER-8B   & \underline{69.94} & \underline{95.77} \\
            \midrule
            
            \multirow{3}{*}{CHARMAP}
            & GPT-4o  & 69.35 & \underline{96.92} \\
             & LLaMA-3.1-8B  & 65.92 & 95.30 \\
             & CoSER-8B   & \underline{70.14} & 95.97 \\
            \midrule

            \multirow{3}{*}{DREAM}
            & GPT-4o  & 69.97 & \underline{\textbf{97.53}} \\
             & LLaMA-3.1-8B  & 65.29 & 95.18 \\
             & CoSER-8B   & \underline{\textbf{71.25}} & 96.03 \\
            \midrule
            
        \end{tabular}
\end{table}

\section{Experiments}
In this section, we comprehensively evaluate DREAM for role-playing ability using the LLM-as-a-judge paradigm~\citep{zheng2023judging, li2025generation}, human evaluation, and objective multiple-choice tasks.

\subsection{Evaluation Metrics}
Following CoSER~\citep{wang2025coser}, we evaluate simulated role-play conversations using GPT-4o as a critic across four key dimensions: (1) \textbf{Storyline Consistency}, (2) \textbf{Anthropomorphism}, (3) \textbf{Character Fidelity}, and (4) \textbf{Storyline Quality}. Assesses the naturalness of simulated conversations, focusing on narrative flow and logical consistency. Detailed rubrics are provided in Appendix~\ref{sec:appendix_metrics}.

For the LIFECHOICE benchmark, we report decision-making accuracy in RPAs on multiple-choice behavioral selections and focus on two key aspects: (1) decision accuracy based on given context and problem (2) decision accuracy with given motivations.

To supplement the indicators lacking in the existing evaluations, we propose TCM benchmark and focus on two key aspects: 
(1) coherent temporal memory, and 
(2) long-term causal narrative consistency. 

We define two metrics to further assess simulated role-playing dialogues: 
(1) \textbf{Future Knowledge Leakage:} Assesses whether dialogues avoid future leakage with respect to the current timeline.
(2) \textbf{Causal Consistency:} Assesses whether RPAs behaviors follow causal logic and are grounded in the character's cognition and experiences. 

\begin{table}[t!]
    \caption{The win rate (\%) from the TCM evaluation, Comparing DREAM Based on Memory Retrieval with RAG, GraphRAG and GCA simulation method from CoSER, where DM. refers to DREAM, FKL. refers to Future Knowledge Leakage and CC. refers to Causal Consistency.}
    \label{tab:tcm_result}
        \begin{tabular}{llcc}
            \toprule
            \textbf{Model} & \textbf{Method} & \textbf{FKL.} & \textbf{CC.} \\
            \midrule
            
            \rowcolor{gray!15}
            \multicolumn{4}{c}{\textbf{Closed-Source LLMs}} \\
            \midrule
            
            \multirow{3}{*}{GPT-4o} 
            & DM. vs RAG    & 95.37 & 79.30 \\
             & vs GraphRAG  & 91.43 & 72.59 \\
             & vs GCA       & 82.19 & 58.12 \\
            \midrule
            
            \multirow{3}{*}{Gemini-2.5-Pro}
            & DM. vs RAG    & 93.51 & 85.26 \\
             & vs GraphRAG  & 90.45 & 78.91 \\
             & vs GCA       & 85.19 & 63.32 \\
            \midrule
            
            \rowcolor{gray!15} 
            \multicolumn{4}{c}{\textbf{Open-Source LLMs}} \\ 
            \midrule
            
            \multirow{3}{*}{LLaMA-3.1-8B}
            & DM. vs RAG    & 78.45 & 78.67 \\
             & vs GraphRAG  & 80.77 & 75.83 \\
             & vs GCA       & 61.55 & 67.35 \\
            \midrule
            
            \multirow{3}{*}{Qwen-2.5-72B}
            & DM. vs RAG    & 82.54 & 75.85 \\
             & vs GraphRAG  & 88.32 & 69.38 \\
             & vs GCA       & 78.86 & 59.31 \\
            \midrule
    
            \rowcolor{gray!15} 
            \multicolumn{4}{c}{\textbf{Role-Playing LLMs}} \\ 
            \midrule
            
            \multirow{3}{*}{Character-GLM-6B}
            & DM. vs RAG    & 95.67 & 82.35 \\
             & vs GraphRAG  & 93.16 & 78.59 \\
             & vs GCA       & 85.83 & 69.95 \\
            \midrule
            
            \multirow{3}{*}{CoSER-8B}
            & DM. vs RAG    & 91.45 & 72.67 \\
             & vs GraphRAG  & 88.67 & 66.90 \\
             & vs GCA       & 89.35 & 56.33 \\
            \bottomrule
        \end{tabular}
\end{table}

\subsection{Experimental Settings} \label{sec:EXP_setting}
\textbf{Baselines} We compare DREAM against four representative baselines: (1) \textbf{RAG} ~\citep{lewis2020retrieval}: Retrieves text chunks most semantically similar to the scenario and incorporates them into role-playing prompts for LLM response generation. (2) \textbf{GraphRAG}~\citep{edge2024local}: Retrieves nodes and community-level summaries most relevant to the scenario from an indexed entity–relationship knowledge graph, and simulates role-playing by combining task descriptions with retrieved information as prompts. (3) \textbf{Given-Circumstance Acting (GCA)}~\citep{wang2025coser}: Utilizes LLMs to construct a book-based global static profile containing rich character information, including persona, plots, experiences, and motivations, and simulates conversations based on the profile and scenario. (4) \textbf{CHARMAP}~\citep{xu2025character}: Retrieves scenario-specific memories via vector matching and combines them with character profiles for persona-driven decision-making.

\textbf{Dataset} We evaluate DREAM on three datasets: 
1) \textbf{CoSER}~\citep{wang2025coser}: We randomly select 100 books from the test set to evaluate general role-playing ability via dialogue simulation; 
2) \textbf{LIFECHOICE}~\citep{xu2025character}: A dataset focusing on RPAs' decision-making ability in multiple-choice questions, where we randomly select 100 samples for evaluation; 
3) \textbf{TCM}: We identify the top 100 books on Goodreads' Best Books Ever list\footnote{\url{https://www.goodreads.com/list/show/1.Best_Books_Ever}}, and select 20 books covering various genres as the evaluation dataset for TCM. By constructing specific scenarios that can detect future knowledge leakage and causal narrative ability, to test the coherent temporal memory and long-term causal narrative consistency of RPAs.

\textbf{Models} Our experiments cover numerous LLMs: 1) Closed-source models, including GPT-4o~\citep{hurst2024gpt} and Gemini-2.5-Pro~\citep{comanici2025gemini}; 
2) Open-source models, including LLaMA-3.1-8B-Instruct~\citep{grattafiori2024llama} and Qwen-2.5-72B-Instruct~\citep{Yang2024Qwen25TR}; 
and 3) Role-playing models, including CoSER-8B~\citep{wang2025coser} and CharacterGLM-6B~\citep{zhou2024characterglm}.

\textbf{Implementation Details}
 For each benchmark, following the method outlined in \S~\ref{sec:graph_construction}, we construct EMG with an average of 32 events and 4,087 nodes per graph, and generate 30 challenging evaluation scenarios for each book. We use LLM-as-a-judge to conduct pairwise comparisons of dialogues generated by different methods across TCM metrics~\citep{chen2024hollmwood}. We construct an independent EMG for each test role in benchmarks. Implementation Details see Appendix~\ref{sec:impl_detail}.

All LLM-based evaluations are conducted using GPT-4o as the judge. We verified the credibility of LLM-as-a-judge through human evaluation by randomly selecting 60 scenarios from the 600 TCM test cases. Results are presented in Appendix~\ref{sec:human_eval_consistency}.

\subsection{Main Results}
\paragraph{Results on CoSER Benchmark} 
We evaluate DREAM alongside baseline models on the CoSER benchmark to assess general role-playing capabilities. Results, averaged over 10 runs, are presented in Table~\ref{tab:coser_results}. Benefiting from event-based character profiles and structured knowledge captured in memory, DREAM consistently outperforms baseline models in storyline consistency, achieving a 14.50\% improvement and significantly reducing semantic drift in long-term interactions. With the Event Memory Graph (EMG) featuring causal narratives, it not only ensures that each response is consistent with the character in terms of style but also logically based on the causal cognitive chain. Therefore, DREAM excels in storyline quality, achieving a 14.88\% improvement and effectively guaranteeing the story generation ability of LLMs in role-playing tasks. In addition, by leveraging dynamic character profiles, DREAM is almost on par with professional role-playing models in terms of anthropomorphism and character fidelity.

\paragraph{Results on LIFECHOICE Benchmark} 
We evaluate DREAM and other methods on LIFECHOICE for RPAs based on multi-choice questions. As shown in Table~\ref{tab:lifechoice_results}, on the professional role-playing model CoSER, the accuracy score of DREAM has increased by approximately 1.21 points, and on the general model GPT-4o, DREAM has also achieved an improvement of approximately 1.82 points. Benefiting from the causal chain and ability to prevent future knowledge leakage brought by DREAM, the model can better alleviate the problems of hallucinations and semantic drift in role-playing. It is worth noting that even when adding the motivations provided by the Benchmark, DREAM still maintains a stable performance advantage over closed-source and open-source models, which reflects the robustness and generalization ability of this method in role-playing and memory-related tasks.

\paragraph{Results on TCM Benchmark} 
Table~\ref{tab:tcm_result} presents the winning rates of DREAM compared to the baselines in TCM evaluation. Compared with RAG and GraphRAG, DREAM maintains a significant advantage in terms of Future Knowledge Leakage capability because it adds timestamps to character memories, with a win rate of over 90\% on Closed-Source and Role-Playing LLMs. Even when limited by the relatively small parameter size of Open-Source LLMs, DREAM still achieves a win rate of over 80\%. Moreover, even a relatively comprehensive role-playing method like GCA, in the absence of EMG, performs far worse than DREAM. In terms of Causal Consistency capability, DREAM achieves an overall win rate of approximately 72\%. Therefore, it can be seen that incorporating a dynamically retrieved causal cognitive chain into the role-playing process can provide a basis for character behaviors in long-term and evolving role-playing, avoid superficial style imitation, and significantly reduce the problem of model hallucinations.

\begin{table}[t!] 
    \caption{Ablation study results (average scores) on CoSER Test. w/o T.S., w/o M.R. and w/o D.P. refer to removing timestamp, memory retrieval and dynamic profile, respectively.}
    \label{tab:ablation}
    \setlength{\tabcolsep}{3pt}
        \begin{tabular}{lcccc}
            \toprule
            \textbf{Model} & \textbf{Complete} & \textbf{w/o T.S.} & \textbf{w/o M.R.} & \textbf{w/o D.P.}\\
            \midrule
            GPT-4o          & 62.86 & 58.18 & 57.67 & 59.37\\
            Gemini-2.5-Pro  & 59.44 & 57.03 & 57.33 & 57.16\\
            LLaMA-3.1-8B    & 57.22 & 54.95 & 54.16 & 55.34\\
            Qwen-2.5-72B    & 59.76 & 58.17 & 57.45 & 58.05\\
            CharacterGLM-6B & 53.79 & 53.15 & 53.23 & 52.94\\
            CoSER-8B        & 58.06 & 56.97 & 56.35 & 57.17\\
            \midrule
        \end{tabular}%
\end{table}

\subsection{Ablation Study}We conducted ablation studies on the main features of DREAM as shown in Table~\ref{tab:ablation}. That is, we design three variants of DREAM by removing timestamp(w/o T.S.), dynamic profile(w/o D.P.) and the memory retrieval capability(w/o M.R.). This comparison was conducted on the CoSER benchmark, aiming to measure the role-playing ability after the lack of key capabilities.

Memory Retrieval (M.R.) has the largest impact. Removing it leads to the most substantial performance drop, approximately 8.9\%, highlighting the central role of the constructed EMG in supporting role-playing capabilities. Notably, for high-capacity models like GPT-4o, the score plummets from $62.86$ to $57.67$. This underscores that the ability to retrieve causally relevant anchor events is indispensable for maintaining narrative depth in complex role-playing.

Timestamps (T.S.): The integration of timestamps is essential for preventing future knowledge leakage (spoilers). Without T.S., models consistently show a performance decline. This suggests that while LLMs possess strong reasoning capabilities, they require explicit temporal constraints to correctly order retrieved memories.

Dynamic Profile(D.P.) ensures that character cognitions and motivations evolve alongside the narrative. Its removal results in a reduction about 5.4\% in character consistency.

In summary, the synergy between the EMG-based retrieval, temporal grounding, and dynamic profiling allows DREAM to achieve superior role-playing fidelity that cannot be matched by any single component in isolation.

\section{Conclusions}
This paper presents DREAM, a novel framework for role-playing that grounds dynamic character profiles in temporally and causally structured event-centric memory. DREAM transforms static literary texts into Event-Aware Memory Graph (EMG), from which dual-granularity retrievable character profiles are generated and updated via memory retrieval for event-aware role-playing. In addition, we introduce the Temporal Causal Memory (TCM) benchmark to evaluate RPAs’ temporal consistency, memory accuracy, and causal narrative ability, complementing existing role-playing evaluations. Experiments show that DREAM, without task-specific fine-tuning, achieves state-of-the-art results among all evaluated baselines across our proposed TCM benchmark and two existing RPA benchmarks, outperforming existing role-playing models. Our approach demonstrates the effectiveness of structured memory in enhancing the interpretability and fidelity of role-playing agents. Future work will explore memory mechanisms that incrementally integrate interaction-generated dialogues into the EMG, enabling agents to evolve continuously.

\section*{Limitations}
Despite DREAM’s strong performance on CoSER and TCM benchmarks, it has limitations:
Role-playing fidelity depends on extraction precision—EMG preserves core narrative logic but may abstract fine-grained stylistic details, leaving room for more expressive future modeling.
Though memory construction is a one-time offline process, graph-based multi-hop retrieval and dynamic profile generation incur higher inference latency and token costs than standard vector-based RAG.
Consistent with common benchmarking standards, we use LLM-based judges~\citep{zheng2023judging, li2025generation}. While aligned with human judgment, these metrics are limited by automated assessment inherent traits, a standard consideration in current narrative evaluation.

\section*{Ethics Statement}
In this paper, we introduce the DREAM framework. The design and utilization of this framework are strictly guided by ethical principles to ensure its applications yield beneficial outcomes for society. We have curated and extracted sample data from a diverse and representative set of literary novels and plays from both domestic and international sources; these texts are utilized exclusively for the construction and evaluation of natural language processing models, with the objective of advancing scientific research in this field. While our framework is designed to enhance role-playing systems, dialogue generation still retains a degree of unpredictability. Consequently, in high-stakes or sensitive scenarios, it is essential to perform rigorous safety scrutiny on generated responses to ensure their appropriateness. We encourage the responsible use of DREAM for educational, entertainment, and creative purposes, while discouraging any harmful or malicious activities.

\clearpage
\bibliographystyle{ACM-Reference-Format}
\bibliography{DREAM}


\appendix
\section{Prompts and KG Definitions for Macro and Micro-Level Extraction}
\label{sec:KG_appendix}
In this section, we provide the detailed ontology definitions and extraction prompts used in our Dual-Granularity Knowledge Extraction module\S~\ref{sec:info_extract}.

\subsection{Knowledge Graph Schema Definition}
To capture both the global context and fine-grained narrative dynamics, we designed two distinct schemas for the Event-Aware Memory Graph (EMG).

\paragraph{Macro-Level Event Context.}
As mentioned in Section \S~\ref{sec:info_extract}, we define 8 types of entities and 7 types of relations to capture the static context of an event $E_i$. Table~\ref{tab:macro_def} details the definitions of these schema elements, covering world settings, background information, and character attributes.

\paragraph{Micro-Level Dynamic Narrative.}
To model the causal evolution within $U_p$, we define 8 types of entities and 12 types of relations. As shown in Table~\ref{tab:micro_def}, these definitions focus on the Emotion-Cognition-Behavior chains, explicitly representing the logical flow of character development.

\section{Implementation Details}
\label{sec:impl_detail}
In CoSER, we replaced the original profile with the DREAM profile generated based on the scenario. In LIFECHOICE and TCM, we use the complete DREAM method, which includes a dynamic profile and memory retrieval.

We maintain the number of dialogue turns and scenarios consistent across all methods. Each experiment involves 2–4 simulated scenarios, and each scenario-method pair is run 10 times. The highest and lowest scores are discarded, and the mean of the remaining runs is taken as the final score. 

\subsection{Extraction Prompts}
We employ Large Language Models (LLMs) to extract structured information from raw literary texts. The prompts are designed to ensure consistency in entity naming and coreference resolution. Table~\ref{tab:macro_prompt} and Table~\ref{tab:prompt_micro} present the specific instructions used for Macro-Level and Micro-Level extraction, respectively.

\section{Detailed Evaluation Metrics}
\label{sec:appendix_metrics}

Following the methodology of CoSER~\citep{wang2025coser}, we employ GPT-4o as a critic to evaluate simulated role-playing conversations. The detailed definitions for the four key dimensions are as follows: 

(1) \textbf{Storyline Consistency:} Assesses alignment between simulated conversations and original dialogue, focusing on whether RPAs’ responses (emotions, attitudes, behaviors) remain faithful to the narrative context. 

(2) \textbf{Anthropomorphism:} Assesses if RPAs act human-like via rubrics covering self-identity, emotional depth, persona coherence, and social interaction. 

(3) \textbf{Character Fidelity:} Assesses how well RPAs reflect character, including linguistic style, knowledge and background, personality, behavior, and relationships. 

(4) \textbf{Storyline Quality:} Assesses the naturalness of simulated conversations, focusing on narrative flow and logical consistency.

\section{Consistency with Human Evaluation}
\label{sec:human_eval_consistency}

To validate the reliability of our model-based evaluation approach used in the TCM benchmark (\S~\ref{sec:EXP_setting}), we conducted a comprehensive agreement analysis between the LLM judge (GPT-4o) and human assessments.

\paragraph{Setup.}
We recruited 5 human annotators who are avid readers and ``fans'' of the corresponding literary works to ensure they possess the necessary domain knowledge to judge character behavior and plot consistency. 
As mentioned in Section (\S~\ref{sec:EXP_setting}), we randomly sampled 60 evaluation scenarios from the total 600 test scenarios in the TCM benchmark. For each scenario, annotators were presented with pairwise outputs (DREAM vs. Baseline) and asked to determine the winner (or tie) across two dimensions: \textbf{Future Knowledge Leakage (FKL.)} and \textbf{Causal Consistency (CC.)}.

\paragraph{Agreement Analysis.}
To quantify the inter-rater reliability between the model's judgments and human annotations, we utilized Cohen's Kappa coefficient ($\kappa$), which accounts for the possibility of chance agreement.
The results are presented in Table~\ref{tab:human_kappa}. The analysis reveals Cohen's Kappa scores ranging from 0.68 to 0.79 across the three metrics.
Specifically, \textbf{FKL.} achieved the highest agreement ($\kappa=0.786$), likely because temporal contradictions (e.g., mentioning future events) are objectively verifiable.
\textbf{CC.} also demonstrated substantial agreement ($\kappa > 0.65$).
These results indicate a high level of consistency between human and machine evaluations, confirming that our LLM-as-a-judge paradigm is sufficiently reliable for reflecting the actual performance differences in the TCM benchmark.

\begin{table}[t]
\begin{tabular}{lcc}
\toprule
\textbf{Metric} & \textbf{FKL.} & \textbf{CC.}\\
\midrule
\textbf{Cohen's Kappa ($\kappa$)} & 0.786 & 0.688\\
\bottomrule
\end{tabular}
\caption{The Cohen's Kappa ($\kappa$) agreement between Human Evaluation and LLM-based Evaluation on TCM metrics.}
\label{tab:human_kappa}
\end{table}

\section{Prompts Demostration}
\label{sec:prompts}
We provide the details of the prompt templates of DREAM in this section.

The prompt for incremental character profile updating and refinement is displayed in Table~\ref{tab:prompt_profile_update}. The prompt for character-centric event recognition and classification is displayed in Table~\ref{tab:prompt_event_recognition}.

\begin{table*}[h]
\caption{Entity and Relation Definitions for Macro-Level Event Context extraction.}
\label{tab:macro_def}
\begin{tabularx}{\textwidth}{l|l|X}
\toprule
\textbf{Category} & \textbf{Name} & \textbf{Description} \\
\midrule
\multirow{8}{*}{Entity} 
 & World Setting & The worldview settings of the current event. \\
 & Environment & The description and characteristics of the environmental scene. \\
 & Dialogue & The dialogue or inner monologue of the core character. \\
 & Relation & The relationship between the selected character and other characters. \\
 & Identity & The identity of the selected character. \\
 & Character & Other characters related to the core character in the event. \\
 & Appearance & The description of the selected character's appearance. \\
 & Action & Specific actions or habits of the selected character. \\
\midrule
\multirow{7}{*}{Relation} 
 & has\_worldview & Links an event to its corresponding World Setting. \\
 & has\_env & Links an event to its environmental description. \\
 & has\_relation\_with & Relationships between characters. \\
 & has\_identity & Describes the role or identity of selected character in the specific event. \\
 & has\_dialogue & Links a dialogue to selected character. \\
 & has\_action & Links a specific action to selected character. \\
 & has\_appearance & Links the appearance description to the selected character. \\
\bottomrule
\end{tabularx}
\end{table*}

\begin{table*}[h]
\centering
\small
\caption{Entity and Relation Definitions for Micro-Level Dynamic Narrative extraction.}
\label{tab:micro_def}
\begin{tabularx}{\textwidth}{l|l|X}
\toprule
\textbf{Category} & \textbf{Name} & \textbf{Description} \\
\midrule
\multirow{8}{*}{Entity} 
 & UnitPlot & Sub-plot units that constitute the selected characteristics of an event. \\
 & Character & Other characters related to the selected character in the event. \\
 & Emotion & Specific emotions of the selected character in the Sub-plot. \\
 & Cognition & Cognitive views formed/updated by the selected character in the Sub-plot. \\
 & Behavior & Key behavioral actions in the Sub-plot that promote the development of the event. \\
 & Scene & Specific scene/location where the event occurs. \\
 & Item & Interactive and owned objects of the selected character's behaviors. \\
 & Skill & Special skills or abilities of the selected character. \\
\midrule
\multirow{12}{*}{Relation} 
 & next\_plot & Defines the temporal sequence between unit-plots. \\
 & core\_role & The core character of the sub-plot. \\
 & other\_role & Other participants in the Sub-plot. \\
 & has\_behavior & Core behaviors of {Role} in the Sub-plot. \\
 & interact\_object & Interactive objects (item or skill) in the behavior. \\
 & occurs\_at & Association with the scene where the plot occurs. \\
 & emotion\_from & Triggering cause of character's emotion. \\
 & cognition\_from & Source of formation of character's cognition. \\
 & cognition\_update\_to & Changes in character's cognition. \\
 & motivated\_by & Driving factors of character's behavior. \\
 & prefer\_to & Specific objects of character's likes and preferences. \\
 & prefer\_cause & Reasons for preferences. \\
\bottomrule
\end{tabularx}
\end{table*}

\begin{table*}[h]
\centering
\small
\caption{Prompt for Macro-Level Event Context Extraction.}
\label{tab:macro_prompt}
\begin{tabularx}{\textwidth}{X} 
\hline
\textbf{\# Your Role: An assistant adept at mining information based on the event content and requirements of the book ``\{BookName\}'' provided by users} \newline
\newline
\textbf{\# Task Requirements} \newline
1. Users will provide the content text corresponding to the events for extraction. \newline
2. Complete the following tasks and conduct extraction as richly and meaningfully as possible. \newline
3. The use of personal pronouns such as ``you'', ``I'', ``he'', ``she'' is prohibited; descriptions must use the clear and unified character names mentioned in the event summary. \newline
\newline
\textbf{\# Task 1: Extract Content by Predefined Categories} \newline
1. WorldSetting: The worldview settings of the current event \newline
2. Environment: The description and characteristics of the environmental scene of the current event \newline
3. Dialogue: The `Dialogue or Inner monologue' of the core character \{Role\}, as well as the corresponding linguistic style (tone, commonly used vocabulary, expression habits) \newline
4. Relation: The relationship between the core character \{Role\} and other characters (e.g., friend, foe, leader, follower, etc.) \newline
5. Identity: The identity of the core character (titles, nicknames, aliases, and other designations that can refer to \{Role\}) \newline
6. Character: Other characters related to the core character in the event \newline
7. Appearance: The description of the core character \{Role\}'s appearance in the event \newline
8. Action: Specific actions or habits of the core character \{Role\} that can reflect the character's personality, emotions, persona, representativeness, and iconicity (e.g., pushing up glasses when thinking, preferring to speak with a soft chuckle when acting as a deity, speaking with a sigh when helpless, being patient to explain when elaborating, habitually roaring when angry......) \newline
\newline
\textbf{\# Important Notes:} \newline
- Add as rich and meaningful attributes as possible during the content extraction process to facilitate the subsequent construction of knowledge graph triples for the event. \newline
- If there is no corresponding content for a category during extraction, ignore that category. \newline
- Extract entities in the order they appear in the text. \newline
\newline
\textbf{\# Initial Settings} \newline
Your role positioning: An assistant adept at extracting information based on the event content and requirements of the book ``\{BookName\}'' provided by users. Strictly comply with the above specific task requirements and output results in \{Language\}. \\
\hline
\end{tabularx}
\end{table*}

\begin{table*}[h]
\centering
\small
\caption{Prompt for Micro-Level Dynamic Narrative Extraction.}
\label{tab:prompt_micro}
\begin{tabularx}{\textwidth}{X} 
\hline
\textbf{\# Your role: Event Knowledge Graph Constructor based on the content of the book \{BookName\}} 
\newline
Based on the complete event text, with the specified core character \{Role\} as the center, systematically sort out the development context of Sub-plot units, emotional dynamic changes, cognitive iteration processes, and behavioral logical associations in the event content.
Finally, generate an Event Knowledge Graph adapted to role-playing scenarios, providing a basis for the Role Memory Module and Behavioral Decision-making. \newline
\newline
\textbf{\# Core Task Objectives} \newline
1. With the core character \{Role\} as the center, disassemble the large event content into logically connected Sub-plot units to form a traceable event chain. \newline
2. Summarize the emotional fluctuations [including triggering reasons and lasting impacts], cognitive changes [including cognitive sources and subsequent effects] of the core character \{Role\} in the Sub-plots, and associate the physical elements of the events [behaviors, scenes, objects]. \newline
3. The constructed graph must fit the needs of role-playing, which can not only restore the original appearance of the event but also support the role to replicate the corresponding emotional state, follow cognitive logic, and echo past events in subsequent interactions. \newline
4. Prohibit the use of pronouns such as you, I, he, she, etc. Must use the clear role names mentioned in the event summary for description, for example: \{Role\}. \newline
5. The final output is in the JSON format of the example \newline
\newline
\textbf{\# Ontology Definition} \newline
\textbf{**1. Nodes [Entities]:**} \newline
1. UnitPlot: Sub-plot units that constitute the core characteristics of an event \newline
2. Character: All roles in the event, the core character is fixed as \{Role\}, ``other\_role'' are labeled with specific role names \newline
3. Emotion: Specific emotions of the core character in the Sub-plot \newline
4. Cognition: Cognitive views formed/updated by the core character in the Sub-plot [need to label the type of cognition: Basic Cognition/Value Judgment/Behavioral Norm] \newline
5. Behavior: Key behavioral actions of the core character in the Sub-plot that promote the development of the event \newline
6. Scene: Specific scene/location where the event occurs \newline
7. Item: Interactive and owned objects of the core character \{Role\}'s behaviors \newline
8. Skill: Special skills and abilities of the core character \{Role\} \newline
\newline
\textbf{**2. Relationships [Edges]:**} \newline
1. next\_plot: Temporal sequence association between small plots [``UnitPlot1'', ``next\_event'', ``UnitPlot2'']. \newline
2. core\_role: Core leader \{Role\} of the Sub-plot [``UnitPlot'', ``core\_role'', ``Role''] \newline
3. other\_role: Other participants in the Sub-plot [``UnitPlot'', ``other\_role'', ``Role''] \newline
4. has\_behavior: Core behaviors of \{Role\} in the Sub-plot [``UnitPlot'', ``has\_behavior'', ``Behavior''] \newline
5. interact\_object: Interactive objects in the behavior [``UnitPlot'', ``interact\_object'', ``Object''] \newline
6. occurs\_at: Association with the scene where the plot occurs [``UnitPlot'', ``occurs\_at'', ``Scene''] \newline
7. emotion\_from: Triggering cause of \{Role\}'s emotion [``Emotion'', ``emotion\_from'', ``Summary for forming the emotion''] \newline
8. cognition\_from: Source of formation of \{Role\}'s cognition [``Cognition'', ``cognition\_from'', ``Summary of the reasons for forming the cognition''] \newline
9. cognition\_update\_to: Changes in \{Role\}'s cognition [``old\_cognition'', ``cognition\_update\_to'', ``new\_cognition''] \newline
10. motivated\_by: Driving factors of \{Role\}'s behavior [``Behavior'', ``motivated\_by'', ``Emotion/Cognition''] \newline
11. prefer\_to: Specific objects of \{Role\}'s likes and preferences [``Sub-plot X'', ``prefer\_to'', ``Object''] \newline
12. prefer\_cause: Reasons for preferences [``Object'', ``prefer\_cause'', ``Cognition''] \newline
\newline
\textbf{\# Task Execution Requirements} \newline
1. Input Adaptation: Based on the large event content provided by the user, conduct structured disassembly within the existing event framework. \newline
2. Core Anchoring: The sorting out of all Sub-plots, emotions, and cognition must revolve around \{Role\}, prioritizing the presentation of the event perception and psychological changes from their perspective. \newline
3. Format Specification: Directly output the structured JSON list of ``unit\_events'' in the reference example, without wrapping the final output content with ```json. \newline
4. Strictly abide by the above specific task requirements, and output a structured json list in \{Language\}, which conforms to the json structure in the example. \newline
\newline
\textbf{\# User Input Text Example: \{User\_input\_Example\}} \newline
\textbf{\# Example of structured output in final json format:\{out\_put\_example\}} \\
\hline
\end{tabularx}
\end{table*}

\begin{table*}[t]
\centering
\small
\caption{Prompt for Event Recognition and Classification.}
\label{tab:prompt_event_recognition}
\begin{tabularx}{\textwidth}{X} 
\hline
\textbf{\# Your Role: An event understanding assistant proficient in the book \{BookName\}, who combines the user-provided single chapter content and previous plot summaries} \newline
\newline
\textbf{\#\# Task 1: Summarize Content Type} \newline
1. Define the candidate output types as: [``complete event'', ``incomplete event'', ``non-character event'', ``contextual content''] \newline
2. Definition scope and explanation of types: \newline
\quad - A ``complete event'' refers to the content of the chapter that involves event of the role \{Role\}, and the content of the current chapter completely describes the event. \newline
\quad - An ``incomplete event'' refers to an event involving the character \{Role\}, but the chapter content provided by the user is insufficient to fully describe the current event, or the current event has not ended and requires additional content from subsequent chapters. \newline
\quad - A ``non-character event'' refers to an event in the current chapter content that involves other characters, does not directly include the character \{Role\}, but the chapter content fully describes the event. \newline
\quad - A ``contextual content'' refers to content in the current chapter that is not an event, such as background descriptions, previous plot summaries, introductions of character relationships, and other similar contextual content descriptions. \newline
3. Based on the specific content of the chapter in \{BookName\} provided by the user and whether \{Role\} is the central figure, output a content type. \newline
4. The summarized content type must strictly follow the defined array. \newline
5. NOTE: Task 1 is only a preliminary task for Task 2 and Task 3, and should not be output separately in the end. \newline
\newline
\textbf{\#\# Task 2: Determine Content Type} \newline
1. If the content type output in [Task 1] is [``complete event''], do not output it first and continue to [Task 3] before outputting. \newline
2. If the content type output in [Task 1] is [``incomplete event''], output strictly in accordance with the example array format and content: [\{\{``event\_type'':``incomplete event'', ``event\_time'':``time marker'', ``event\_name'':``summarized name of the incomplete event'', ``event\_description'':``summary of the content of the incomplete event chapter''\}\}], and end all tasks. \newline
3. If the content type output in [Task 1] is [``non-character event''], output strictly in accordance with the example array format and content: [\{\{``event\_type'':``non-character event'', ``event\_time'':``time marker'', ``event\_name'':``summarized name of the non-character event'', ``event\_description'':``summary of the content of the non-character event chapter''\}\}], and end all tasks. \newline
4. If the content type output in [Task 1] is [``contextual content''], output strictly in accordance with the example array format and content: [\{\{``event\_type'':``contextual content'', ``event\_time'':``time marker'', ``event\_name'':``summarized name of the contextual content'', ``event\_description'':``summary of the content of the contextual content chapter''\}\}], and end all tasks. \newline
\newline
\textbf{\#\# Task 3: Summarize and Extract Events and Requirements} \newline
1. Take \{Role\} as the central figure of the event content. \newline
2. Based on the specific content of the chapter in \{BookName\} provided by the user, summarize it into an event centered on \{Role\}. \newline
3. According to the chapter content, summarize a time or time period that can mark the sequence of events in the book; if the time cannot be determined, use the chapter number as the time marker. \newline
4. It is forbidden to use personal pronouns such as you, I, he, she, etc. The event summary must use clear character names, character titles, or character identities, etc. \newline
5. The output content and format are: [\{\{``event\_type'':``complete event'', ``event\_time'':``time marker'', ``event\_name'':``summarized event name'', ``event\_description'':``summarized and induced event content description''\}\}] \newline
\newline
\textbf{\#\# Initialization} \newline
You act as: An event understanding assistant proficient in the book \{BookName\}, who refers to previous plot summaries and combines the user-provided single chapter content. Strictly follow the order of tasks and output the results using \{Language\}. \\
\hline
\end{tabularx}
\end{table*}

\begin{table*}[h]
\centering
\small
\caption{Prompt for Incremental Character Profile Updating and Refinement.}
\label{tab:prompt_profile_update}
\begin{tabularx}{\textwidth}{X} 
\hline
\textbf{Role: You are an expert literary analyst and character profiler of the book \{BookName\}. } \newline
Your task is to \textbf{continuously update and refine} the Character Profile for \{Role\} as the narrative progresses. \newline
\newline
\textbf{Inputs:} \newline
1. \textbf{Existing Profile:} The character's profile derived from previous events (if any). \newline
2. \textbf{New Event Data:} Extracted knowledge graph data from the current event \{Timeline: T\_current\}, including Identity, Background, Description, Style, Traits, Motivations, Relationships, Causal Narrative and Cognition. \newline
\newline
\textbf{Instructions:} \newline
1. Incremental Update: DO NOT discard the ``Existing Profile''. Use the ``New Event Data'' to enrich, verify, or evolve the existing profile. \newline
\quad - \textbf{Reinforce:} If new data confirms existing traits, strengthen the description. \newline
\quad - \textbf{Add:} If new data reveals previously unknown aspects (e.g., a new relationship or hidden skill), add them to the relevant section. \newline
\quad - \textbf{Evolve:} If the character undergoes a change (e.g., from calm to angry, or a change in worldview), explicitly describe this evolution in the ``Cognitive State'' or ``Key Experiences'' section. \newline
\quad - \textbf{Contextualize:} Resolve conflicts based on the timeline. If the character was ``Loyal'' in T1 but ``Betrayed'' in T2, the profile should reflect this shift. \newline
2. Synthesize: Write in coherent, literary paragraphs. Do not simply append lists. Merge new information naturally into the existing structure. \newline
3. Output Format: Strictly follow the section structure below. Output the results using \{Language\}. \newline
\newline
\textbf{Target Structure:} \newline
\textbf{**Name:**} \{Role\} \newline
\textbf{**Background:**} (Update based on new status or revealed backstory) \newline
\textbf{**Appearance:**} (Add details if appearance changes or new features are described) \newline
\textbf{**Linguistic styles:**} (Update if the character's tone shifts in this event) \newline
\textbf{**Personality Traits:**} (Refine traits based on new actions) \newline
\textbf{**Core Motivations:**} (Update current drives and underlying motives) \newline
\textbf{**Relationships:**} (Update dynamic relations with others) \newline
\textbf{**Cognition Chains:**} (CRITICAL: Summarize the character's state of mind in this event and how it compares to the past) \newline
\textbf{**Key Experiences:**} (Briefly add the essence of the current event to their history) \newline
\textbf{**Causal Narrative Chains**} (Generate a causal narrative paragraph description based on provided triples) \\
\hline
\end{tabularx}
\end{table*}
\end{document}